%% file: main.tex
\documentclass[11pt]{article}

\usepackage[]{acl}

\usepackage{times}
\usepackage{latexsym}
\usepackage{combelow}
\usepackage{xcolor}
\usepackage{graphicx}
\usepackage{pifont}
\usepackage{adjustbox}
\usepackage{booktabs}
\usepackage{multicol}
\usepackage{multirow}
\usepackage{floatrow}
\usepackage{svg}

\definecolor{tokencolor}{HTML}{BB3E00}
\definecolor{charcolor}{HTML}{F7AD45}

\usepackage{fullpage}
\usepackage{tikz}

\usepackage{tcolorbox}
\tcbuselibrary{minted,breakable,xparse,skins}

\newcommand{\checkmark}{{\color{teal} \ding{52}}}
\newcommand{\xmark}{{\color{red} \ding{55}}}

\usepackage{algorithm}
\usepackage{hyperref}
\usepackage{algpseudocode}

\usepackage{amssymb,amsmath,amsthm}

\usepackage[T1]{fontenc}
\usepackage[utf8]{inputenc}
\usepackage{microtype}

\title{The Strawberry Problem:\\Emergence of Character-level Understanding in Tokenized Language Models}

\author{Adrian Cosma\textsuperscript{1,2}, \cb{S}tefan Ru\cb{s}e\cb{t}i\textsuperscript{1}, Emilian R\u{a}doi\textsuperscript{1}, Mihai Dasc\u{a}lu\textsuperscript{1} \\
    {\textsuperscript{1}National University of Science and Technology POLITEHNICA Bucharest} \\
    {\textsuperscript{2}Dalle Molle Institute for Artificial Intelligence Research (IDSIA)} \\
    {\small\texttt{\{ioan\_adrian.cosma, stefan.ruseti, emilian.radoi, mihai.dascalu\}@upb.ro}}}

\begin{document}
\maketitle
\begin{abstract}
Despite their remarkable progress across diverse domains, Large Language Models (LLMs) consistently fail at simple character-level tasks, such as counting letters in words, due to a fundamental limitation: tokenization. In this work, we frame this limitation as a problem of low mutual information and analyze it in terms of concept emergence. Using a suite of 19 synthetic tasks that isolate character-level reasoning in a controlled setting, we show that such capabilities emerge suddenly and only late in training. We find that percolation-based models of concept emergence explain these patterns, suggesting that learning character composition is not fundamentally different from learning commonsense knowledge. To address this bottleneck, we propose a lightweight architectural modification that significantly improves character-level reasoning while preserving the inductive advantages of subword models. Together, our results bridge low-level perceptual gaps in tokenized LMs and provide a principled framework for understanding and mitigating their structural blind spots. We make our code publicly available.
\end{abstract}

\section{Introduction}
\label{sec:intro}
\input{sections/1.intro}

\section{Related Work}
\label{sec:related}
\input{sections/2.related}

\section{Method}
\label{sec:method}
\input{sections/3.method}

\section{Experiments \& Results}
\label{sec:exp}
\input{sections/4.experiments-results}

\section{Conclusions}
\label{sec:conc}
\input{sections/5.conclusions}

\section*{Limitations}
\label{sec:lim}
\input{sections/6.limitations}

\section*{Acknowledgments}
\label{sec:ack}
\input{sections/7.ack}

\bibliography{refs}

\appendix
\section{Appendix}
\label{sec:appendix}

\input{tables/vocab-maker}
\input{tables/task-maker}

\end{document}

%% file: sections/1.intro.tex
LLMs have exhibited impressive capabilities in solving olympiad math problems \cite{trinh2024solving}, playing open-world games \cite{zihao2023mincraft} and passing bar exams \cite{achiam2023gpt}. However, paradoxically, LLMs often struggle in simple tasks that involve character-level reasoning and manipulation\footnote{This can be seen as a form of Moravec's Paradox \cite{newell1983intellectual} - reasoning is easy; perception is hard.}. A growing body of work shows that language models are brittle to misspellings, struggle with character-level tasks \cite{shin2024large,zhang2024large}, and fail even simple reasoning tasks that require access to words' constituent letters. One such infamous problem, the so-called "Strawberry Problem", consists of counting the number of "r"s in the word "strawberry", a problem that many foundational models struggle to consistently answer even today \cite{chai-etal-2024-tokenization}.

\begin{figure}[t!]
    \centering
    \includegraphics[width=1.0\linewidth]{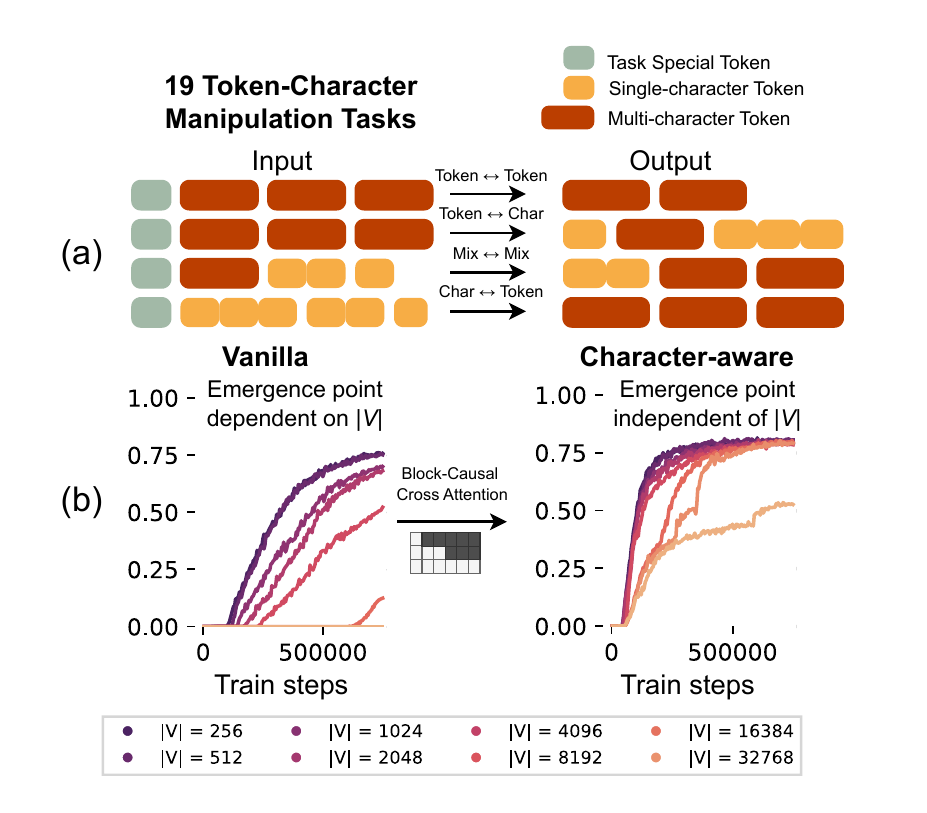}
    \caption{Tokenization obscures character composition of words. (a) We develop 19 token manipulation tasks spanning token$\leftrightarrow$token, token$\leftrightarrow$char, char$\leftrightarrow$token, and mixed settings, isolating character reasoning from semantics. (b) In a standard transformer decoder LM, character-level understanding emerges abruptly and late in training, having the emergence point pushed back as the tokenizer vocabulary size $|V|$ increases; adding our character-aware module brings early emergence that is effectively independent of $|V|$.}
    \label{fig:teaser}
\end{figure}


The root of this problem lies in text tokenization. Tokenization is heavily used in modern language models \cite{sennrich2015neural}, in which the raw text is compressed into sequences of multi-character subword tokens. This comes at a cost: tokenization severs the connection between words and their characters, limiting the model's reasoning capabilities about characters and morphology. Paradoxically, while tokenization imposes a structural bottleneck, it also provides critical inductive biases \cite{rajaraman2024toward}, and cannot be completely avoided. In the absence of tokenization, models trained directly on characters or bytes (e.g., \citet{xue-etal-2022-byt5,wang2024mambabyte}) learn more slowly and require more data to generalize. Thus, there is a fundamental tension: tokenization improves efficiency and generalization at the cost of losing fine-grained perceptual access to the underlying text \cite{chai-etal-2024-tokenization}.


In this work, we argue that the emergence\footnote{We consider the definition by \citet{more-is-different}: \textit{"Emergence is when quantitative changes in a system result in qualitative changes in behavior"}. This applies not only to model size, but, for example, to increasing training duration.} of character-level understanding is best modeled through the lens of mutual information and theories of concept emergence \cite{lubana2025a}. Our thesis is that learning character composition of words is equivalent to learning underreported commonsense facts \cite{do2024really}. Human-written text almost never directly mentions the characters inside words, as this is self-evident for humans upon reading a text\footnote{A phenomenon that can be understood as a form of non-reporting bias \cite{gordon2013reporting,shwartz2020neural}, in which the rare and the interesting are overrepresented at the expense of the trivial.}. 

To better understand this phenomenon, we construct a suite of 19 synthetic tasks that require models to reason about the character composition of tokens in a strictly controlled, synthetic, setting (see Figure \ref{fig:teaser}). We show that performance on these tasks emerges late in training, is modulated by vocabulary size and composition, and aligns with theoretical predictions from concept percolation models of emergence \cite{lubana2025a}. A model trained on tokenized sequences gets little signal about characters and must slowly reconstruct this mapping across many training steps, and we show and that real world tokenizers worsen this effect.

We introduce a simple architectural intervention to address this bottleneck: a block cross-attention mechanism that exposes character-level information to the model alongside token embeddings. Unlike existing byte-level or hybrid models \citep{tay2022charformer,neitemeier2025hierarchical}, our approach preserves the inductive benefits of subword tokenization while mitigating its perceptual blindness. We show that this design improves character-level reasoning with minimal additional cost, and that it effectively raises the mutual information between tokens and characters during training. 
Our experiments are reproducible and we make our code publicly available\footnote{\href{https://github.com/cosmaadrian/strawberry-problem}{https://github.com/cosmaadrian/strawberry-problem}}.

Our work makes the following contributions:

\begin{enumerate}
    \item We develop a benchmark of 19 synthetic tasks to train and evaluate character-level understanding in tokenized LMs, revealing slow and sudden emergence patterns across training. We show that character learning is slow and dependent on vocabulary size and number of characters per token, even in an idealized setting, with the effect being heightened using real-world data.
    
    \item We find that we can explain the emergence of character-level understanding through concept percolation theories of emergence \cite{lubana2025a}. Our results hint that learning character-token correspondences is not fundamentally different from learning abstract concept-property associations.
    
    \item We propose a lightweight character-aware architecture that increases the empirical mutual information between tokens and their characters. Our design adds a small cross-attention module that allows each token to attend to its constituent characters, while still using tokens as inputs and outputs. We validate our architecture on models using pretrained tokenizers and on real-world data, indicating significant improvements in character-level tasks compared to a standard language model operating on tokenized sequences.
\end{enumerate}

%% file: sections/2.related.tex
\paragraph{Internal Character Representation in Language Models.}
Recent works \cite{shin2024large,zhang2024large,chai-etal-2024-tokenization} highlight the limitations of Language Models in understanding the character-level structure of tokens. \citet{shin2024large} argued that LLMs lack robust internal representations of the character composition of tokens. In discussing future directions, they propose embedding tokens with character information and positional encodings - an approach closely aligned with our method. \citet{zhang2024large} evaluated LLMs on 15 simple text-editing tasks and found that models struggle without fine-tuning; the authors found that supervised fine-tuning for text editing substantially improved performance without harming general capabilities. \citet{kaplan2025from} argued that LLMs implicitly combine subword units into full words and exploited this finding to improve efficiency by adding dedicated word-level tokens. In contrast, our focus is the reverse: we investigate how LLMs can be encouraged to decompose tokens into their constituent characters. Different from these previous works \cite{shin2024large,zhang2024large}, we isolate and analyze character-level capabilities in a tightly controlled synthetic setting, revealing their emergence dynamics during training. 

\paragraph{Character-aware models.} 
There have been multiple works attempting to design character-aware models, such as the works of \citet{tay2022charformer,islam2022vocabulary,wang2024mambabyte} \textit{et alia}, by operating directly on characters, bypassing the need for tokenization. One downside of such models is that they model directly the input and output characters, resulting in long generation sequences and decreased efficiency \cite{rajaraman2024toward}. In contrast, our model operates directly on multi-character tokens, utilizing the inductive bias given by tokenization, while incorporating character information for each token.

The construction of neural architectures with a hierarchical structure of representations has been a common design pattern, usually in domains that require either long contexts \cite{he2024hdt,nawrot2021hierarchical,wu-etal-2021-hi} or high-detail granularity \cite{chen2021crossvit}. In contrast to previous works, we design our model for causal next-token prediction in mind, and not for MLM or for computing general representations.

\paragraph{Theories of capability emergence.} 
Our work is also related to recent theoretical perspectives on capability emergence in models \cite{mckenzie2023inverse,hupkes2020compositionality,lubana2025a,park2024emergence}. \citet{hupkes2020compositionality} designed controlled tasks to test models' capability to compositionally generalize. While they operated at the level of tokens, we explore whether similar emergent behaviours arise at the lower level of token decomposition into characters. \citet{mckenzie2023inverse} identified tasks in which larger LLMs perform worse than their smaller counterparts. Among these tasks is the "resisting correction" task, in which the model automatically, but wrongly, corrects a misspelled token. However, the study's focus was on model scale and compute allocation, while we explore the relationship between vocabulary size and performance. \citet{lubana2025a} proposed a framework for studying emergent capabilities using context-sensitive grammars and compositional tasks, under the theory of bipartite graph percolation. While our scope and domain are different, we show that the framework of graph percolation still applies and can explain the learning dynamics observed in our setup, hinting that learning token-character correspondence is a similar problem to learning correspondences between concepts and their properties.

%% file: sections/3.method.tex
\subsection{Tokenization-Induced Information Bottleneck}
Let $\Sigma$ be a character alphabet and let $C=(c_1,\dots,c_n)\in\Sigma^n$ denote the character sequence (spelling) of a word. Let $W$ denote the lexical identity (word type) determined by $C$: there is a deterministic map $g:\Sigma^*\to\mathcal{W}$ with $W = g(C)$. Let $X$ be the corpus context, all surrounding tokens excluding the word at the current position. The quantity of interest is the empirical mutual information \cite{shannon1998mathematical}:

\begin{align}
    \hat{I}(X; C) = \mathbb{E}\left[\log\frac{p(x,c)}{p(x)\,p(c)}\right]
\end{align}

Namely, how much the context $X$ tells us about the characters $C$ of the word that appears in that context. Because $W$ is a deterministic function of $C$, we have $I(W;C) = H(C) = H(W)$. Moreover, natural text exhibits a certain phenomenon: humans do not need to explicitly mention the characters in a word, so the context $X$ provides little signal about $C$, which means the empirical mutual information $\hat{I}(X; C)$ is low, even though the theoretical mutual information $I(W; C)$ is high, since $W$ deterministically determines $C$ (i.e., $ \hat{I}(X; C) \ll I(W; C)$). Thus, for a language model trained only on tokenized word sequences, the empirical mutual information $\hat{I}(X; C) \approx 0$, unless the model is explicitly character-aware. Similar to the presence of commonsense facts, this data sparsity reflects a form of reporting bias \cite{shwartz2020neural}: humans do not encode character-level details in natural text, leading to under-representation of this information in the training signal. However, learning commonsense knowledge requires explicit data collection \cite{speer2017conceptnet}, whereas learning character-token correspondences is comparatively simpler: we know at all times which characters comprise a word, and we can leverage this in the design of a character-aware architecture.

\subsection{Experimental Setup}

\input{tables/tasks}

\paragraph{Word-level and Character-level tasks.} 
Previous works explored pretrained LLMs' performance on several character-level tasks \cite{shin2024large,zhang2024large} and show that their performance is sub-par. In this work, we create a set of 7 word-level tasks and 12 character-level tasks to systematically explore emergent capabilities for character manipulation across training. Compared to previous works, our tasks are easier and do not involve counting or multi-hop reasoning (e.g., count vowels of every even word). Table \ref{tab:tasks} shows our tasks alongside examples for parameters, inputs, and desired outputs. By design, the tasks have input-output combinations tokenized either as words $\leftrightarrow$ words (all word-level tasks), characters $\leftrightarrow$ words (dirty-input character tasks), words $\leftrightarrow$ characters (clean-input character tasks), or a mix of tokenizations (e.g., "Rewrite uppercase" / "Replace letters"). As such, for character-level tasks, multi-character tokens might be imperfectly split into characters (e.g., "Remove letter"), and models are forced to indirectly learn token-character correspondence across many training steps. Tasks are evaluated using an exact match between the model output and the desired output. While using exact match metrics impacts evaluation curves \cite{schaeffer2023are}, they are correlated with other softer metrics such as log-probabilities, and inflection points between memorization and generalization phases match between the two \cite{lubana2025a}. Furthermore, an exact match enables us to compare performance unambiguously across different tokenizers and vocabulary sizes.

\paragraph{Vocabulary construction.}
We opted for a strictly controlled and synthetic experimental environment to test the capability of tokenized language models to learn character-level tasks and to eliminate as many confounding factors as possible. We generate a fixed-length vocabulary of words $V$, which is comprised of all single-character letters, including uppercase, numbers, and a space character. Multi-character tokens are all comprised of the same number of characters, $K$, uniformly sampled (see Appendix \ref{sec:appendix}: Table \ref{tab:vocab-maker}). In our work, $K \in \{4, 6, 8\}$ and $|V| \in \{2^8, \dots, 2^{15}\}$. To encode a task, we use a special task token for each task, which is optionally followed by parameters (see Table \ref{tab:tasks}). Consequently, our tokenizer is comprised of single characters, numbers, and multi-character words, each with its unique ID. As such, if a multi-character word is corrupted, it will be represented through individual characters as a fallback tokenization, with no intermediate subwords.

\begin{figure*}
    \centering
    \includesvg[width=0.95\linewidth]{images/strawberry-diagram.drawio.svg}
    \caption{Diagram of our character-aware language model. During inference, each token attends to its corresponding characters using a block-causal cross-attention operation. Characters are encoded alongside their positions within their corresponding tokens using a small 1-block Transformer decoder, using a block-causal self-attention mechanism. MLPs are omitted in the figure for brevity.}
    \label{fig:diagram}
\end{figure*}

For our analysis, we ignore language grammar, since none of the tasks require grammar manipulation, only token and character manipulation. To construct sentences, words are sampled uniformly from the tokenizer vocabulary (see Appendix \ref{sec:appendix}: Table \ref{tab:task-maker}), ignoring the Zipfian distribution of real-world languages \cite{Piantadosi2014-fr}. However, we also test performance on randomly sampled sentences from Wikipedia, using two pretrained tokenizers (i.e., GPT-2 \cite{radford2019language} and LLaMA-2 \cite{touvron2023llama2}). It is expected that real-world texts would not qualitatively change learning dynamics; still, they would make learning harder due to the imbalanced distribution of characters per word and different word lengths.

In this most simplified version of the problem, the only factor influencing model performance is its ability to connect tokens with their characters, which appear fragmented and inconsistent across training. Such a strictly controlled environment is similar to other concurrent works \cite{allen2023physics} aiming to explain model capabilities without real-world confounders. In this setup, the model is forced to learn the algorithm behind each task, through so-called "induction heads" \cite{olsson2022context} or "name mover heads" \cite{wang2022interpretability}, since the tasks only require token-level manipulations and not semantic understanding \cite{shin2024large}.

\subsection{Generating tokens by attending to characters.}

We design a lightweight character-aware module that complements the main Transformer decoder to increase the mutual information between tokens and their constitutive characters. In our design, we were guided by several criteria: \textit{(i)} the character-aware module must be lightweight \textit{(ii)} the model output type must remain unchanged (i.e., still output multi-character tokens), \textit{(iii)} there is an unambiguous correspondence between tokens and their characters, and \textit{(iv)} there is an unambiguous order of characters inside a token.

Figure \ref{fig:diagram} showcases our architecture. Given these criteria, we designed a small, 1-layer Transformer block that uses a \textit{Block-Causal Self-Attention} mask to process characters. Since the main model operates on multi-character tokens, whenever a new token is generated, the character model has access to all its characters, removing the need for a casual diagonal attention matrix. The block-causal attention mask enables the module to attend to all characters in the current token, as well as previous characters from previous tokens, but does not "cheat" by attending to the characters of future tokens. The order of characters inside a token is encoded using learnable \textit{Intra-Token Position} embeddings, similar to Abacus Embeddings \cite{mcleish2024abacus}. The dimensionality of the character encoder can be made smaller than the main module (in our case, $d_{chars} = \frac{1}{2}d_{tokens} = 256$). We also experiment with smaller dimensionalities for the character module (i.e., $d_{chars} \in \{64, 128, 256\}$, corresponding to ratios $\frac{d_{chars}}{d_{tokens}} \in \{\frac{1}{4}, \frac{1}{8}, \frac{1}{16}\}$). After encoding characters, the resulting embeddings interact with the token embeddings through a \textit{Block-Causal Cross-Attention} operation at each layer of the main model. We also experiment with adding the character embeddings at a single layer of the main model, at different positions. The cross-attention operation prevents tokens from attending to future characters and ensures that each token attends to its corresponding characters alongside characters from previous tokens. Character-token correspondence is ensured through learnable \textit{Inter-Token Position} embeddings.

\paragraph{Overview.} The model is efficient by operating on multi-character tokens and not directly predicting characters, leveraging the tokenizer compression, and has explicit knowledge of the character composition of each token. In principle, this design can be extended hierarchically, for example, having tokens attend to their constituent subwords and each subword attending to its constituent characters. While the character module is significantly smaller than the main module, it still suffers from the quadratic complexity of the attention operation. Presumably, the character encoder can be made more efficient to avoid quadratic attention by utilizing, for example, local attention patterns \cite{beltagy2020Longformer} or by using more specialized modules such as linear recurrent units \cite{orvieto2023resurrecting}. Our model is reminiscent of other works in computer vision, such as CrossViT \cite{chen2021crossvit}, and is part of a larger pattern of designing architectures that use hierarchical representations \cite{nawrot2021hierarchical,chalkidis2022exploration,he2024hdt}. Nevertheless, this pattern is more common in computer vision than in NLP. This hierarchical character-to-token cross-attention design addresses the problem of "perception" of current LLMs, which capture high-level semantic meaning, but struggle at "high-resolution", in terms of perceiving individual characters of each token.

\subsection{Training configuration}

All models were trained on a uniform sample over all tasks for 750k iterations. We used Adam \cite{kingma2014adam} optimizer, a batch size of 64, and a learning rate of 0.00001, annealed using a cosine decay scheduler \cite{loshchilov2016sgdr}. The baseline model has 10M parameters, excluding embedding matrices, across 8 layers, with a model dimensionality of 512. Similarly, our model has 11M parameters, with 1M being allocated to the character encoder. The character encoder is a lightweight, single-block Transformer with a dimensionality of 256. One important advantage of our experimental setup is that it is readily reproducible on a single A100 GPU. Training took approximately one day per run for all $\sim$60 runs. Models were trained with "infinite" data, since input sentences and tasks were generated on-the-fly. In the case of training on sentences from Wikipedia, we pre-generated 5M sentences. In our experiments, every hyperparameter is kept fixed, except for the vocabulary and the tokenizer.

\begin{figure*}[t!]
    \centering
    \includesvg[width=0.85\linewidth]{images/emergence-step-merged.drawio.svg}
    \caption{Emergence point for acquisition of character-level \textit{(top)} and word-level \textit{(bottom)} understanding across vocabulary sizes for the "vanilla" model and our character-infused model. For our model, capabilities emerge early on in training for character-level tasks, and do not depend on $|V|$ or $|K|$.}
    \label{fig:emergence-step}
\end{figure*}

\begin{figure}[hbt!]
    \centering
    \includesvg[width=0.95\linewidth]{images/emergence_char_tasks.svg}
    \caption{Evolution of average accuracy over character-level tasks. Using standard transformer decoder \textit{(top)}, the emergence point depends on the vocabulary size and $K$, whereas our architecture \textit{(bottom)} eliminates the differences in emergence points across vocabulary sizes.}
    \label{fig:emergence-chars}
\end{figure}

\begin{figure}[hbt!]
    \centering
    \includesvg[width=0.95\linewidth]{images/emergence_word_tasks.svg}
    \caption{Evolution of average accuracy over word-level tasks. Emergence points for word-level tasks are not affected by vocabulary size as prominently as character-level tasks. While our architecture \textit{(bottom)} targets character-token associations, it still provides improvement by increasing the amount of information per training sample compared to the standard LM \textit{(top)}.}
    \label{fig:emergence-words}
\end{figure}

%% file: tables/tasks.tex
\begin{table*}[hbt!]
    \centering
    \resizebox{1.0\linewidth}{!}{
    \begin{tabular}{clc|l|l}
        & \textbf{Task Name} &  \textbf{Example Param.} & \textbf{Example Input} & \textbf{Example Output}\\
        \midrule
        \multirow{8}{*}{\rotatebox{90}{\textbf{Word-Level}}} & Remove word & red & Strawberries are red and sweet. & Strawberries are and sweet.\\
    & Remove word every K & 2 & Strawberries are red and sweet. & Strawberries red sweet.\\
    & Swap every K words (clean) & 5 & Strawberries are red and sweet. & sweet. are red and Strawberries \\
    & Swap every K words (dirty) & 5 & sweet. are red and Strawberries & Strawberries are red and sweet. \\
    & Replace words & are, red & Strawberries are red and sweet. & Strawberries red red and sweet. \\
    & Reverse the words (clean) & N/A & Strawberries are red and sweet. & sweet. and red are Strawberries \\
    & Reverse the words (dirty) & N/A & sweet. and red are Strawberries & Strawberries are red and sweet. \\
    \midrule
    \multirow{12}{*}{\rotatebox{90}{\textbf{Character-Level}}} & Remove letter & r & Strawberries are red and sweet. & Stawbeies ae ed and sweet. \\
    & Rewrite uppercase every K letters & 3 & Strawberries are red and sweet. & StrAwbErrIes arE rEd And swEet.\\
    & Replace letters & e, s & Strawberries are red and sweet. & Strawbsrriss ars rsd and swsst. \\
    & Rewrite with every K letter & 3 & Strawberries are red and sweet. & Saei eea e.\\
    & Swap every K letters (clean) & 2 & Strawberries are red and sweet. & tSarbwreirsea err dea dns ewte. \\
    & Swap every K letters (dirty) & 2 & tSarbwreirsea err dea dns ewte. & Strawberries are red and sweet. \\
    & Remove letter every K & 4 & Strawberries are red and sweet. & Strwberie ar re an swet. \\
    & Rewrite uppercase every K words & 2 & Strawberries are red and sweet. & STRAWBERRIES are RED and SWEET. \\
    & Reverse words (clean) & N/A & Strawberries are red and sweet. & seirrebwartS era der dna .teews \\
    & Reverse words (dirty) & N/A & seirrebwartS era der dna .teews & Strawberries are red and sweet. \\
    & Reverse (clean) & N/A & Strawberries are red and sweet. & .teews dna der era seirrebwartS \\
    & Reverse (dirty) & N/A & .teews dna der era seirrebwartS & Strawberries are red and sweet. \\
    \end{tabular}
    }
    \caption{Summary of proposed tasks used in our work. The tasks address either word-level or character-level manipulations and optionally require input parameters.}
    \label{tab:tasks}
\end{table*}

%% file: sections/4.experiments-results.tex
In Figure \ref{fig:emergence-chars}, we show the evolution of average accuracy across character tasks for both the base language model and our model that incorporates character information. The emergence point for character understanding tasks is progressively offset as a function of vocabulary size $|V|$ and number of characters per token $K$. In contrast, for our model, the emergence points are stable across $|V|$ and $K$, having reasonable performance gain even in scenarios where the base model has accuracy equal to 0. This effect is also present, although not as prominently, for token understanding tasks (Figure \ref{fig:emergence-words}), since token manipulation tasks can be easily learned by the base model. 

In Figure \ref{fig:emergence-step} we show the emergence step across vocabulary sizes. We plot the training step at which the accuracy for a task is larger than 0.5\%, across training runs of different $|V|$ and $K$. We find that increasing vocabulary size is correlated with a later emergence point for the vanilla model for both word-level and character-level tasks, having a predictable relationship. The addition of the character-aware module eliminates the dependence on $|V|$ for character-level tasks, but to a lesser extent for the word-level tasks.

\subsection{Evaluation on real-world data}

\begin{figure}[hbt!]
    \centering
    \includesvg[width=0.95\linewidth]{images/wiki.svg}
    \caption{The effect of using real sentences sourced from Wikipedia in evaluating character understanding tasks, across two pretrained tokenizers.}
    \label{fig:wiki}
\end{figure}

We trained the baseline language model and our model on sentences sourced from Wikipedia, using two pretrained tokenizers (i.e., GPT-2 \cite{radford2019language} and LLaMa-2 \cite{touvron2023llama2}), with vocabulary sizes of 50K and 32K tokens, respectively. In Figure \ref{fig:wiki} we show our results: incorporating characters has a significant effect on learning dynamics for both tokenizers, with the base model being unable to learn character composition of words across training. As our results point to, this effect will be amplified with larger vocabularies: current LLMs tend to benefit from having progressively larger vocabularies \cite{huang2025overtokenized}, with models such as Gemma 3 \cite{gemmateam2025gemma3technicalreport} operating on a vocabulary of 256K tokens, which also implies more characters per token. Our results indicate that character understanding tasks in tokenized language models are a form of "inverse scaling" \cite{mckenzie2023inverse}: the larger the tokenizer vocabulary, the slower the model learns. 

\subsection{The effect of downsizing the character encoder}

In Figure \ref{fig:intervention}, we show results for varying the position of the cross-attention operation in the main model by incorporating character information either at the beginning, the middle, or the end of the language model. Similarly, we reduce the dimensionality of the character encoder to 12.5\% of that of the main model, making the character model fast and lightweight in terms of memory consumption. Infusing character information in the middle of the model yields the best results, and downsizing the character encoder does not significantly alter the training curves, suggesting that only the presence of characters and their association with tokens is sufficient.

\begin{figure}[hbt!]
    \centering
    \includesvg[width=0.95\linewidth]{images/intervention_layer.svg}
    \caption{The effect of the position of the block-cross attention in the main model \textit{(top)} and that of downsizing the char. encoder \textit{(bottom)}. $|V| = 8192$ and $K = 4$.}
    \label{fig:intervention}
\end{figure}

\subsection{The effect of increasing base model size}

We scaled the model in a principled way using maximal update parametrization ($\mu$P \cite{yang2022tensorprogramsvtuning}) to increase the model width (size of the linear layer) through a multiplicative factor. Using $\mu$P ensures that the hyperparameters for the smaller model (e.g. learning rate) can be directly transferred to larger versions of the same model, making the comparison fair between scales in terms of optimal hyperparameters. We scaled the base encoder, but kept the character encoder fixed. The results presented in Table \ref{tab:model-scale} show that scaling the width of the model reduces the emergence point, but without addition of the character encoder, the acquisition is slow and inefficient. In other words, without the character encoder, model needs $\times$15 more parameters to reach the same performance. Across scales, the size of the character encoder is negligible. These results are for our simplified case, assuming uniform distribution of characters and tokens. Our other results (Figure \ref{fig:wiki}) shows that this effect is greatly amplified with real world data.

\input{tables/model-scale}

\subsection{A percolation model of character understanding.}

In the interest of explaining the offset emergence points for the acquisition character understanding, we applied a percolation model of capability emergence, as described by \citet{lubana2025a}. Readers are referred to the original work for a detailed explanation of this framework, which the authors applied in the context of learning concept-property relationships. In that scenario, emergence points coincided with a critical threshold $p_c$ in which the bipartite graph of concepts and their respective properties ($K$) is fully connected: across training, the model progressively learns edges between concepts and properties until reaching a certain threshold, proportional to $\sqrt{|K|}$, after which the model enters a sudden generalization phase. In our scenario, we have a direct analogy to concept learning: our "concepts" are multi-character tokens, and their "properties" are the set of characters they are composed of. As such, the emergence point should be proportional to the number of edges \cite{newman2001random,cohen2002percolation,lubana2025a}, in our case equaling $\sqrt{|V| * k}$. In Figure \ref{fig:percolation-chars} we show the emergence points for the base language model. Scaling the training steps by $\sqrt{|V| * k}$ results in the collapse of the emergence points. This result indicates no conceptual difference between learning concept-property mappings and learning token-character mappings. 

\begin{figure}[hbt!]
    \centering
    \includesvg[width=0.95\linewidth]{images/percolation_char_tasks.svg}
    \caption{Graph percolation explains emergence points for acquisition of character understanding, similar to concept percolation \cite{lubana2025a}. Scaling training curves by the percolation threshold $\sqrt{|V| * k}$ collapses emergence points across vocabulary sizes.}
    \label{fig:percolation-chars}
\end{figure}

%% file: tables/model-scale.tex
\begin{table*}[hbt!]
    \centering
    \resizebox{0.8\linewidth}{!}{
    \begin{tabular}{ccccp{3cm}p{4.8cm}}
        \textbf{Vocab. Size} & \textbf{Model Width} & \textbf{Base Params} & \textbf{Character-Aware} & \textbf{Params Char. Encoder (\% of base)} & \textbf{Char. Tasks Emergence Step$\downarrow$} \\
        \midrule
        \multirow{6}{*}{8192} & \multirow{2}{*}{512} & \multirow{2}{*}{10M} & \xmark & -- & >750k \\
         & & & \checkmark & 1M (9\%) & \textbf{77k} \\
         \cmidrule{2-6}
         & \multirow{2}{*}{1024} & \multirow{2}{*}{45M} & \xmark & -- & 180k \\
         & & & \checkmark & 1M (2\%) & \textbf{53k} \\
         \cmidrule{2-6}
         & \multirow{2}{*}{2048} & \multirow{2}{*}{150M} & \xmark & -- & 81k \\
         & & & \checkmark & 1M (0.6\%) & \textbf{20k} \\
        \midrule
       \multirow{6}{*}{16384} & \multirow{2}{*}{512} & \multirow{2}{*}{10M} & \xmark & -- & >750k \\
         & & & \checkmark & 1M (9\%) & \textbf{110k} \\
         \cmidrule{2-6}
         & \multirow{2}{*}{1024} & \multirow{2}{*}{45M} & \xmark & -- & 260k \\
         & & & \checkmark & 1M (2\%) & \textbf{53k} \\
         \cmidrule{2-6}
         & \multirow{2}{*}{2048} & \multirow{2}{*}{150M} & \xmark & -- & 110k \\
         & & & \checkmark & 1M (0.6\%) & \textbf{45k} \\
    \end{tabular}
    }
    \caption{The effect of increasing model scale in terms of width. The addition of a small character-aware module substantially reduces the emergence point for the acquisition of character-level understanding. We kept $K = 4$ fixed across all runs.}
    \label{tab:model-scale}
\end{table*}

%% file: sections/5.conclusions.tex
Tokenization is crucial in language modeling, enabling long context and aiding generalization \cite{rajaraman2024toward}. In this paper, we show that for a class of problems that require fine-grained understanding of character composition of tokens, models acquire such information very slowly, predictably dependent on the vocabulary size and number of characters per token. We argued that this is due to non-reporting bias and that this phenomenon is similar to learning commonsense facts from general text. There is a design mismatch in the way in which humans hierarchically perceive written text (from lines, characters, words and phrases) and the way LLMs process text.

To this end, we proposed a lightweight and straightforward architectural modification that eliminates this dependence on vocabulary size and showed that capabilities emerge faster and consistently. Lastly, we applied a theory of capability emergence in concept learning \cite{lubana2025a} and showed that it applied to our setting, equating the phenomena of learning concepts with learning characters' composition in tokens.

%% file: sections/6.limitations.tex
The main limitation of our work is that we conducted most of our experiments in a synthetic and idealized setup to understand the phenomena of character understanding of tokens, without confounding factors. Nonetheless, our proposed architecture showed good results when training on real data, but its impact to real-world scenarios needs to be validated at larger scales.

%% file: sections/7.ack.tex
This research was supported by the project "Romanian Hub for Artificial Intelligence - HRIA", Smart Growth, Digitization and Financial Instruments Program, MySMIS no. 334906.
We also thank Oleg Szehr for his unforgiving feedback that led to a more rigorous presentation of our method.


%% file: tables/vocab-maker.tex
\begin{table}[hbt!]
\definecolor{bg}{gray}{0.95}
\DeclareTCBListing{mintedbox}{O{}m!O{}}{%
  breakable=true,
  listing engine=minted,
  listing only,
  minted language=#2,
  minted style=default,
  minted options={%
    linenos,
    gobble=0,
    breaklines=true,
    breakafter=,,
    fontsize=\scriptsize,
    numbersep=8pt,
    #1},
  boxsep=0pt,
  left skip=0pt,
  right skip=0pt,
  left=20pt,
  right=0pt,
  top=3pt,
  bottom=3pt,
  arc=5pt,
  leftrule=0pt,
  rightrule=0pt,
  bottomrule=2pt,
  toprule=2pt,
  colback=bg,
  colframe=orange!70,
  enhanced,
  overlay={%
    \begin{tcbclipinterior}
    \fill[orange!20!white] (frame.south west) rectangle ([xshift=16pt]frame.north west);
    \end{tcbclipinterior}},
  #3}
\begin{mintedbox}{python}
def make_vocab(vocab_size, K):
    vocab = dict()
    token_id = 0

    numbers = '0123456789'
    letters_lower = 'abcdefghijklmnopqrstuvwxyz'
    letters_upper = letters_lower.upper()
    letters = letters_lower + letters_upper

    # adding the bytes and space
    for token_name in letters + numbers + ' ': 
        vocab[token_name] = token_id
        token_id += 1
    
    # adding multi-character tokens
    for _ in range(vocab_size):
        word = ''.join(random.choice(letters) for _ in range(K))
        vocab[word] = token_id
        token_id += 1
    
    return vocab
\end{mintedbox}
\caption{Python snippet for tokenizer vocabulary creation. In our work, words are made from randomly sampled characters of fixed size $K$. The tokenizer vocabulary contains single-characters, numbers, multi-character words, and a space as individual unique tokens. Additionally, each task is represented as a unique special token.}
\label{tab:vocab-maker}
\end{table}

%% file: tables/task-maker.tex
\begin{table}[hbt!]
\definecolor{bg}{gray}{0.95}
\DeclareTCBListing{mintedbox}{O{}m!O{}}{%
  breakable=true,
  listing engine=minted,
  listing only,
  minted language=#2,
  minted style=default,
  minted options={%
    linenos,
    gobble=0,
    breaklines=true,
    breakafter=,,
    fontsize=\scriptsize,
    numbersep=8pt,
    #1},
  boxsep=0pt,
  left skip=0pt,
  right skip=0pt,
  left=20pt,
  right=0pt,
  top=3pt,
  bottom=3pt,
  arc=5pt,
  leftrule=0pt,
  rightrule=0pt,
  bottomrule=2pt,
  toprule=2pt,
  colback=bg,
  colframe=orange!70,
  enhanced,
  overlay={%
    \begin{tcbclipinterior}
    \fill[orange!20!white] (frame.south west) rectangle ([xshift=16pt]frame.north west);
    \end{tcbclipinterior}},
  #3}
\begin{mintedbox}{python}
def make_task(tokenizer):
    target_task = random.choice(TASKS)
    # choose multi-character words
    word_list = [
        w for w in tokenizer.vocab2id.keys()
        (if len(w) > 1 and w not in SpecialTokens)
    ]
    random_sentence = ' '.join([
        random.choice(word_list) for _ in range(16)
    ])
    task_output = target_task(random_sentence)
    # task_token, param, input, output 
    return task_output 
\end{mintedbox}
\caption{Python snippet for task creation. In our work, a sentence is comprised of 16 uniformly sampled multi-character words from the tokenizer, upon which the task algorithm is enacted.}
\label{tab:task-maker}
\end{table}